\documentclass[sigconf,screen]{acmart}
\usepackage{multirow}
\usepackage{subfigure}
\usepackage{blindtext}
\usepackage{enumitem}
\usepackage{algorithm}  
\usepackage{algpseudocode}  
\usepackage{amsmath}  
\usepackage{hyperref}
\usepackage{flushend}
\usepackage{balance}

\AtBeginDocument{%
\providecommand\BibTeX{{%
  \normalfont B\kern-0.5em{\scshape i\kern-0.25em b}\kern-0.8em\TeX}}}

\copyrightyear{2022} 
\acmYear{2022} 
\setcopyright{acmcopyright}
\acmConference[SIGIR '22]{Proceedings of the 45th International ACM SIGIR Conference on Research and Development in Information Retrieval}{July 11--15, 2022}{Madrid, Spain}
\acmBooktitle{Proceedings of the 45th International ACM SIGIR Conference on Research and Development in Information Retrieval (SIGIR '22), July 11--15, 2022, Madrid, Spain}
\acmPrice{15.00}
\acmDOI{10.1145/xxxxxx.xxxxxx}
\acmISBN{978-1-4503-xxxx-x/xx/xx}

\begin{document}

\title{Deep Page-Level Interest Network in Reinforcement Learning for Ads Allocation}

\author{Guogang Liao$^{1\dagger}$, Xiaowen Shi$^{1\dagger}$, Ze Wang$^{1*}$, Xiaoxu Wu$^1$, Chuheng Zhang$^{2\ddagger}$, Yongkang Wang$^1$, Xingxing Wang$^1$, Dong Wang$^1$}

\affiliation{%
 \institution{$^1$Meituan, Beijing, China \ \ \ \  $^2$IIIS, Tsinghua University, Beijing, China}
 \city{}
 \country{}
}
\email{{liaoguogang, shixiaowen03, wangze18, wuxiaoxu04}@meituan.com, 
zhangchuheng123@live.com,}
\email{{wangyongkang03, wangxingxing04, wangdong07}@meituan.com}

\renewcommand{\shortauthors}{Guogang Liao and Xiaowen Shi, et al.}

\renewcommand{\shortauthors}{Guogang Liao and Xiaowen Shi, et al.}


\begin{abstract}
  \renewcommand{\thefootnote}{\fnsymbol{footnote}}
  \footnotetext[2]{Equal contribution. Listing order is random.}
  \renewcommand{\thefootnote}{\fnsymbol{footnote}}
  \footnotetext[3]{This work was done when Chuheng Zhang was an intern in Meituan.}
  \renewcommand{\thefootnote}{\fnsymbol{footnote}}
  \footnotetext[1]{Corresponding author.}
  \renewcommand{\thefootnote}{\fnsymbol{footnote}}
  A mixed list of ads and organic items is usually displayed in feed and how to allocate the limited slots to maximize the overall revenue is a key problem. Meanwhile, modeling user preference with historical behavior is essential in recommendation and advertising (e.g., CTR prediction and ads allocation). Most previous works for user behavior modeling only model user's historical point-level positive feedback (i.e., click), which neglect the page-level information of feedback and other types of feedback. To this end, we propose Deep Page-level Interest Network (DPIN) to model the page-level user preference and exploit multiple types of feedback. Specifically, we introduce four different types of page-level feedback as input, and capture user preference for item arrangement under different receptive fields through the multi-channel interaction module. Through extensive offline and online experiments on Meituan food delivery platform, we demonstrate that DPIN can effectively model the page-level user preference and increase the revenue for the platform.
  
\end{abstract}

\maketitle

\section{Introduction}
\begin{figure}[tb]
  \centering
  \subfigure[User A focuses on three items in a page]{
  \begin{minipage}[t]{0.5\linewidth}
  \centering
  \includegraphics[height=2.2in]{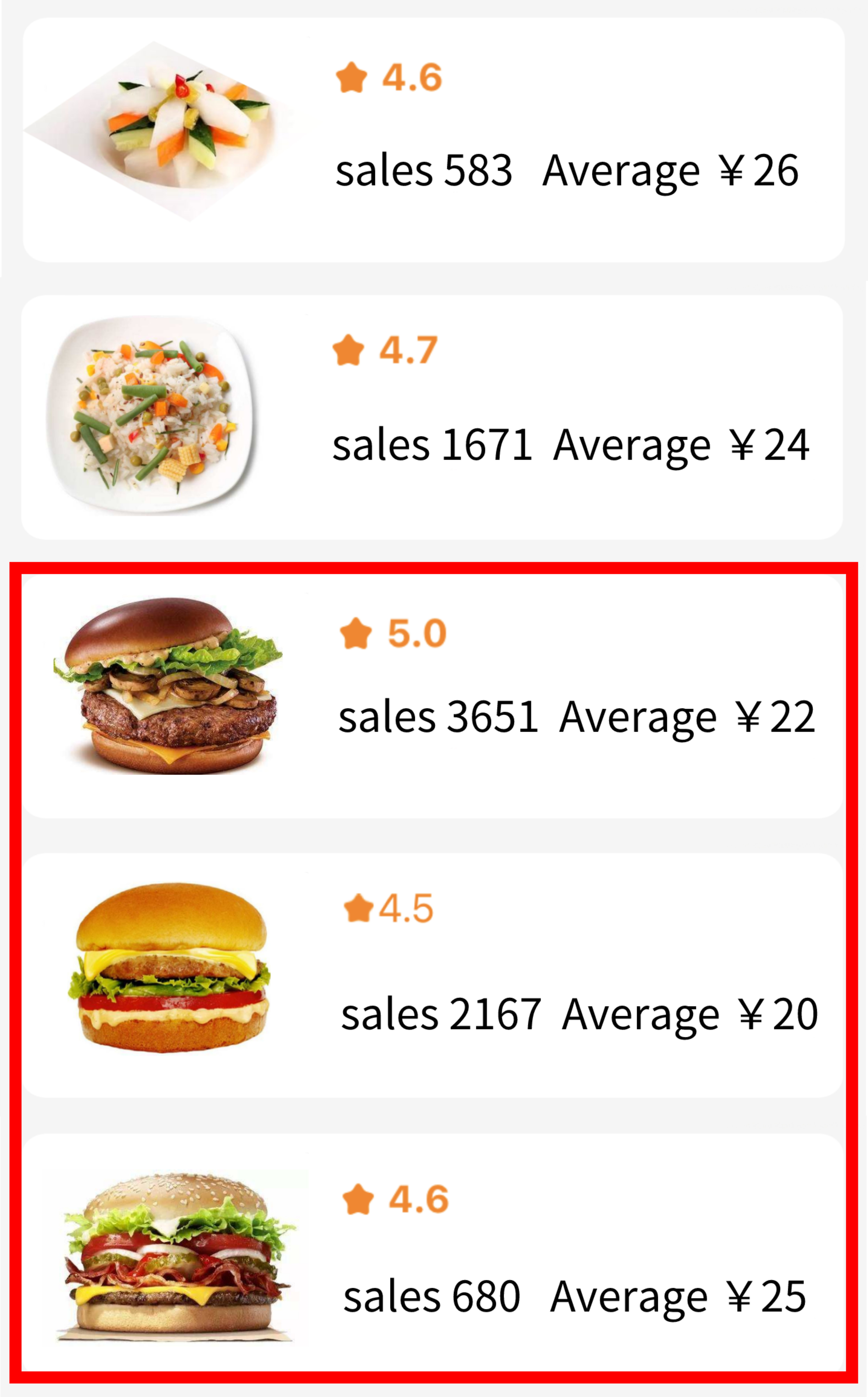}
  \label{fig:subfig:u1} 
  \end{minipage}%
  }%
  \subfigure[User B focuses on two items in a page]{
  \begin{minipage}[t]{0.5\linewidth}
  \centering
  \includegraphics[height=2.2in]{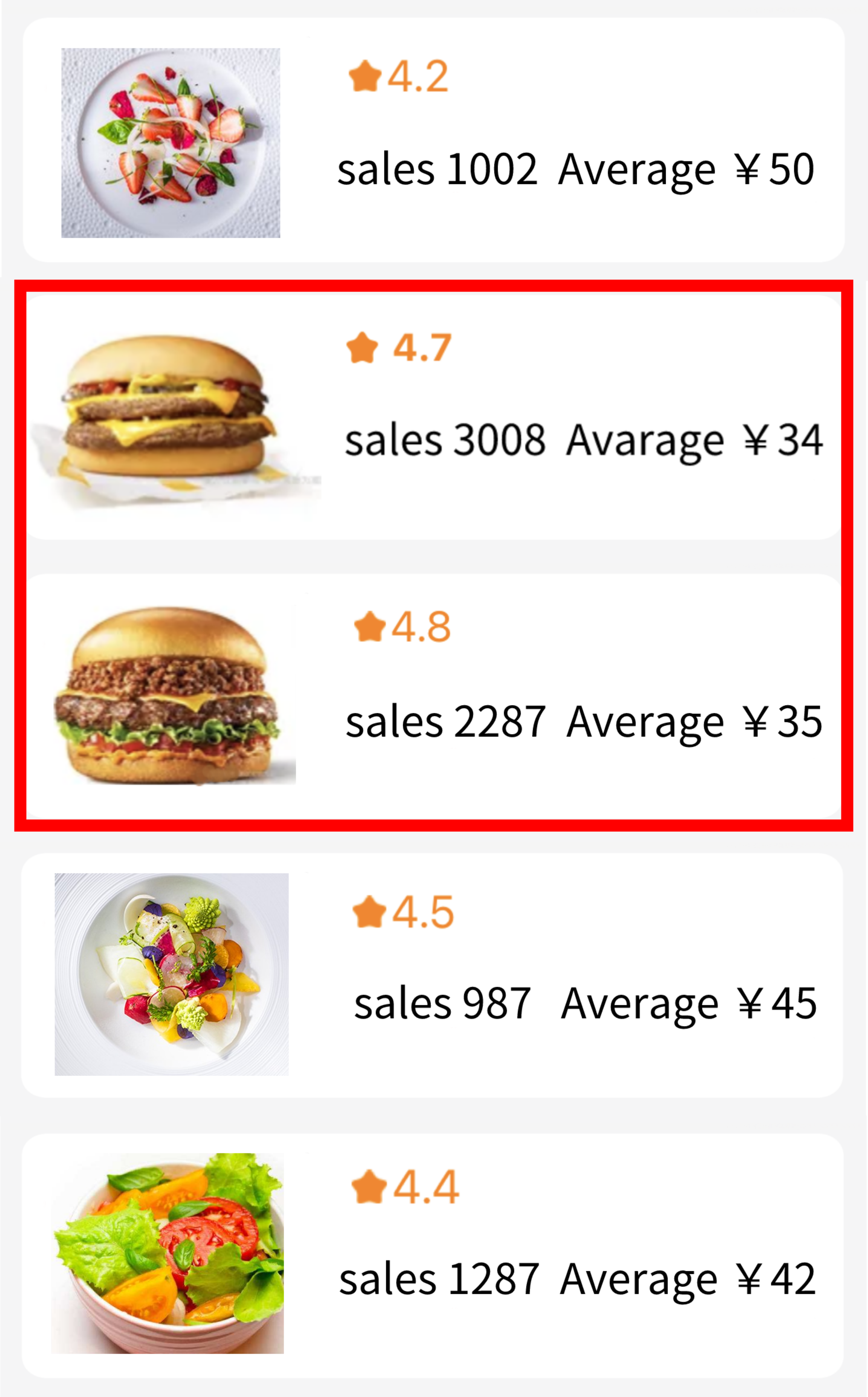}
  \label{fig:subfig:u2} 
  \end{minipage}%
  }%
  \caption{Different users may have different preferences on receptive field when browsing.}
  \label{fig:1}
\end{figure}

Ads and organic items are mixed together and displayed to users in e-commerce feed nowadays \cite{yan2020LinkedInGEA,Ghose2009AnEA,li2020deep} and how to allocate the limited slots to maximize the overall revenue has become a key problem \cite{Wang2011LearningTA,Mehta2013OnlineMA,zhang2018whole}. Since the feed is presented to the user in a sequence, recent ads allocation strategies model the problem as Markov Decision Process (MDP) \cite{sutton1998introduction} and solve it using reinforcement learning (RL) \cite{zhang2018whole,liao2021cross,zhao2019deep, Feng2018LearningTC, zhao2020jointly}. For intance, \citet{xie2021hierarchical} propose a hierarchical RL-based framework to first decide the type of the item to present and then determine the specific item for each slot. \citet{liao2021cross} proposes CrossDQN which takes the crossed state-action pairs as input and allocates the slots in one page at a time.

User behavior modeling, which focuses on learning the intent representation of user interest, is widely introduced in recommendation and advertising scenarios (e.g., CTR prediction and ads allocation) \cite{zhou2019deep, xiao2020deep,pi2019practice}. Most previous works on user behavior modeling \cite{zhou2019deep, xiao2020deep,zhao2021non} only model user interest using positive feedback (e.g., click) while neglect other types of feedback, which may result in an inaccurate approximation of user interest. \citet{xie2021deep} model both positive and negative feedback and achieve better performance. However, they only model point-level feedback, which ignore the page-level information of the feedback (e.g., mutual influence among items in one page). \citet{fan2022modeling} introduce page-wise feedback sequence but still face three major limitations. Firstly, it would be better to
match historical page-level feedback with the target page rather than the target item. Secondly, as shown in Figure \ref{fig:1}, different users may have different preferences on receptive field when browsing, which means users may pay attention to the mutual influence among items within different ranges. Thirdly, they ignore ohter types of page-level feedback (e.g., pull-down, leave).

To address these limitations, we present an method named Deep Page-level Interest Network (DPIN)  to model the page-level user perference for ads allocation and exploit multiple types of feedback. Specifically, we construct four page-level behavior sequence (i.e., page-level order, click, pull-down and leave) and use the Multi-Channel Interaction Module (MCIM) to model page-level user preference. In MCIM, we first use multiple convolution kernels with different sizes to extract the information of different receptive fields on the page. Nextly, we conduct Intra-Page Attention Unit (IPAU) to capture the mutual influence among items within different ranges. Subsequently, we design the Inter-Page Interaction Unit (IPIU) to calculate the correlation between target page and page-level sequences and denoise page-level implicit feedback (i.e., unclick and pull down, hereinafter referred to as pull-down) by sequence interaction.

We have conducted several offline experiments and evaluated our approach on real-world food delivery platform. The experimental results show that the introduction of page-level historical behavior and the modeling of page-level user preference can significantly improve the platform revenue. 
This is a meaningful attempt in modeling page-level user preference on ads allocation.

\section{Problem Formulation}
\label{sec:problem}
In our scenario, items are displayed to the user in the form of page turning. We present $K$ slots in one page and handle the allocation for each page in the feed of a request sequentially.
The ads allocation problem is formulated as a MDP ($\mathcal{S}$, $\mathcal{A}$, $r$, ${P}$, $\gamma$), the elements of which are defined as follows:

\begin{itemize}[leftmargin=*]
  \item \textbf{State space $\mathcal{S}$}. A state $s\in\mathcal{S}$ consists of the candidate items (i.e., the ads sequence and the organic items sequence which are available on current step $t$), the user's base features (e.g., age, gender), the context features (e.g., order time, order location) and four types of user's page-level historical behavior sequences (i.e, page-level order, click, pull-down and leave).
  \item \textbf{Action space $\mathcal{A}$}. An action $a \in \mathcal{A}$ is the decision whether to display an ad on each slot on the current page, which is formulated as follows:
  \begin{equation}
    \begin{aligned}
      &\ \ \ \ \ \ \ \ \ \ \ \ 
      a=(x_{1}, x_{2}, \ldots, x_{K}), \\
      \ \ \ \text{where}\ \ x_{k} = &\begin{cases}
      1& \text{display an ad in the $k$-th slot}\\
      0& \text{otherwise}
    \end{cases},\ \forall k \in [K].\end{aligned}
  \end{equation}
  In our scenario, we do not change the order of the items within ads sequence and organic items sequence. 
  \item \textbf{Reward $r$}. After the system takes an action in one state, a user browses the mixed list and gives a feedback. The reward is calculated based on the feedback and consists of ads revenue $r^\text{ad}$ and service fees $r^\text{fee}$:
  \begin{equation}
    \begin{aligned}
  r(s,a) = r^\text{ad}+r^\text{fee}.\end{aligned}
    \label{eq:eta}
  \end{equation}
  \item \textbf{Transition probability ${P}$}.
  $P(s_{t+1}|s_t,a_t)$ is defined as the state transition probability from $s_t$ to $s_{t+1}$ after taking the action $a_t$, where $t$ is the index for the page.
  When the user pulls down, the state $s_{t}$ transits to the state of next page $s_{t+1}$. 
  The items selected by $a_t$ will be removed from the state on the next step $s_{t+1}$.
  If the user no longer pulls down, the transition terminates. 
  
  \item \textbf{Discount factor $\gamma$}. The discount factor $\gamma \in [0, 1]$ balances the short-term and long-term rewards.
\end{itemize}
  
Given the MDP formulated as above, the objective is to find an ads allocation policy $\pi: \mathcal{S} \rightarrow \mathcal{A}$ to maximize the total reward.
  
\begin{figure*}[tb]
  \centering
  \includegraphics[width=1\linewidth]{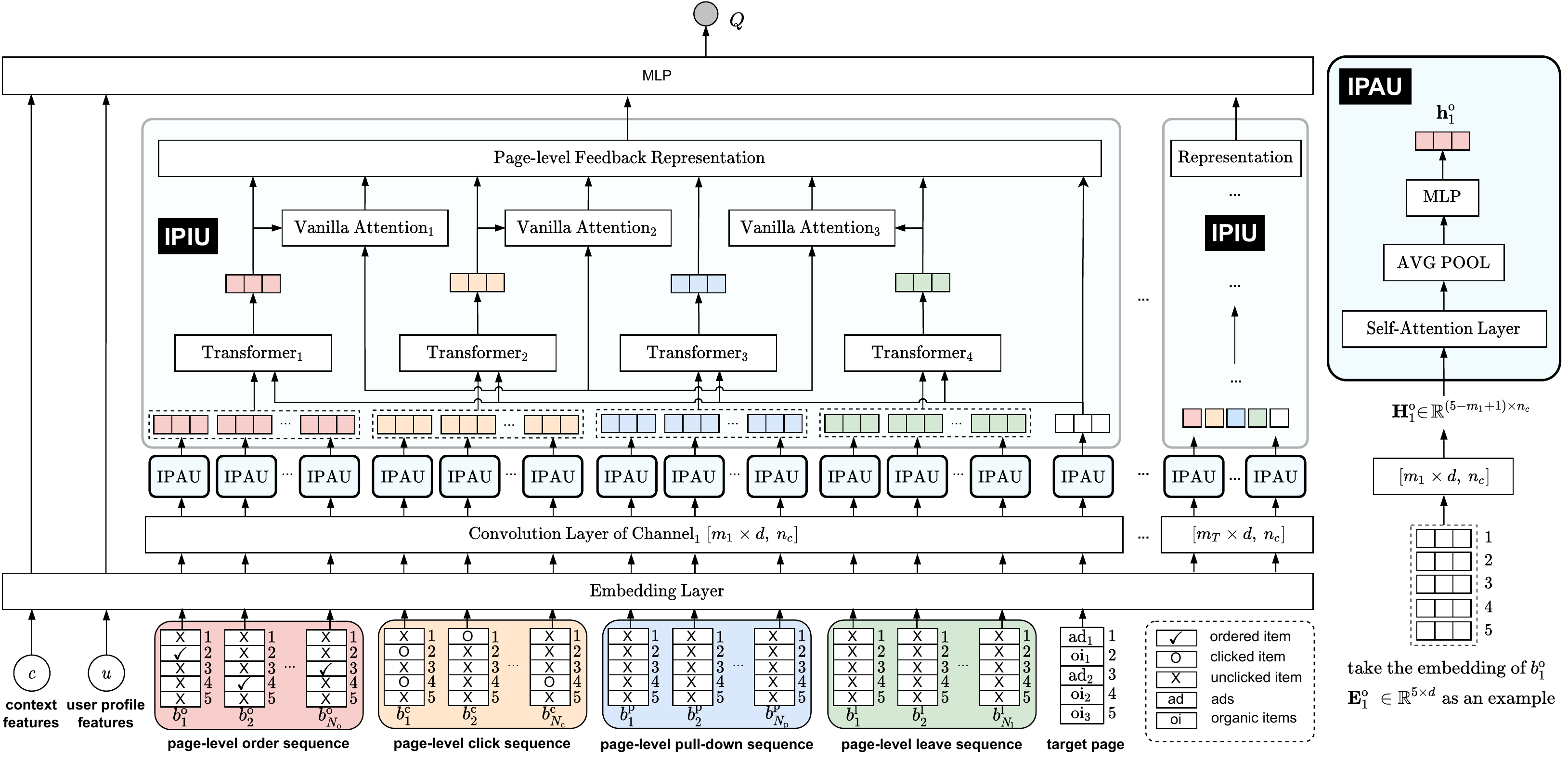}
  \caption{
  The structure of Deep Page-level Interest Network. The features are first input into the Embedding Layer. Then the embeddings of four page-level sequences and target page are input into the multi-channel interaction module to generate representations. The output multiple page-level feedback representations are concatenated with the embeddings of context and user profile to predict the value $Q$.
  }
  \label{fig:fig1}
\end{figure*}

\section{Methodology}
As shown in Figure \ref{fig:fig1}, we first input the features of each sample into the embedding layer to form corresponding embeddings. Nextly, we input the embeddings of four page-level sequences and target page into Multi-Channel Interaction Module (MCIM) to generate $T$ page-level feedback representations. Finally, we concatenate the output representations with the embeddings of context and user profile to predict the value $Q$ through a Multi-Layer Perceptron (MLP). Next, we will detail each part.

\subsection{Input \& Embedding Layer}
The input information consists of five parts: context features, user profile features, four page-level historical behavior sequences, candidate ads and organic items sequences, and candidate actions. 
Similar to \citet{liao2021cross}, we cross the candidate ads and organic items sequences according to the action to form the target page arrangement $t(s,a)$. 
The page-level historical behavior sequences for the user include four types: page-level sequence of order, click, pull-down, and leave. We use "page-level" to refer to the present items on current page when the user's behavior occurred. We concatenate the position feature and feedback category feature for each item to improve the sequence representation.

We use embedding layers to extract the embeddings from raw inputs.
The embedding matrix for page-level information is denoted as $\mathbf{E} \in \mathbb{R}^{K \times d}$, where $K$ is the number of presented items on a page and $d$ is the dimension of embedding.
We denote the embeddings for page-level order sequence, click sequence, pull-down sequence, leave sequence, target page, the user profile, the context as $\{\mathbf{E}_i^\text{o}\}_{i=1}^{N_\text{o}}$, $\{\mathbf{E}_i^\text{c}\}_{i=1}^{N_\text{c}}$, 
$\{\mathbf{E}_i^\text{p}\}_{i=1}^{N_\text{p}}$, 
$\{\mathbf{E}_i^\text{l}\}_{i=1}^{N_\text{l}}$, 
$\mathbf{E}^{t}$, $\mathbf{e}^\text{u}$, and $\mathbf{e}^\text{c}$ respectively,  where the subscript $i$ denotes the index within the sequence and $N_\text{o}$, $N_\text{c}$, $N_\text{p}$ and $N_\text{l}$ are the length of corresponding behavior sequences.

\subsection{Multi-Channel Interaction Module (MCIM)}
Different arrangements of displayed items on a page make different influence on user behaviors. Accordingly, we propose MCIM to model page-level user preferences. Different channels can capture user perference for item arrangement under different receptive fields through three parts: Convolution Layer (CL), Intra-Page Attention Unit (IPAU) and Inter-Page Interaction Unit (IPIU). Next, we will use the structure of a channel to introduce each part.

\subsubsection{Convolution Layer}
Each page-level embedding matrix $\mathbf{E}$ is first input into the convolution layer to extract the local field information of the page:
\begin{equation}
  \begin{aligned}
\mathbf{H}_1 =  f^{m \times d}_{n_c} (\mathbf{E}),
  \end{aligned}
\end{equation}
where $n_c$ is the number of convolution kernels, $m$ is the size of receptive fields, and $\mathbf{H}_1  \in \mathbb{R}^{(K-m+1) \times n_c}$ is the output.

\subsubsection{Intra-Page Attention Unit}
Then the matrix are input into self-attention layer which uses the scaled dot-product attention:
\begin{equation}
  \begin{aligned}
    \mathbf{H}_2 = \text{SDPA}(\mathbf{Q},\mathbf{K},\mathbf{V}) = \text{soft} \max (\frac{\mathbf{Q}\mathbf{K}^\top}{\sqrt{n_c}})\mathbf{V},
  \end{aligned}
\end{equation}
where $\mathbf{Q},\mathbf{K},\mathbf{V}$ represent query, key, and value, respectively. $d$ denotes feature dimension of each feature. Here, query, key and value are transformed linearly from $\mathbf{H}_1$, as follows:
\begin{equation}
  \begin{aligned}
    \mathbf{Q} = \mathbf{H}_1 \mathbf{W}^{Q}, \mathbf{K} = \mathbf{H}_1 \mathbf{W}^{K}, \mathbf{V} = \mathbf{H}_1  \mathbf{W}^{V},
  \end{aligned}
\end{equation}
where $\mathbf{W}^{Q}, \mathbf{W}^{K}, \mathbf{W}^{V} \in \mathbb{R}^{n_c \times n_c}$. Then $\mathbf{H}_2$ are input into a MLP to generate the page-level representation:
\begin{equation}
  \begin{aligned}
    \mathbf{h} = \text{MLP}_1\Big(\text{avg pool}\big(\mathbf{H}_2\big)\Big).
  \end{aligned}
\end{equation}

\subsubsection{Inter-Page Interaction Unit}
\citet{xie2021deep} have proved that historical behaviors which are more relevant to the target item can provide more information for the model's predict. Therefore, we use the multi-head self-attention mechanism to calculate the interactions between target page and page-level sequences. Take the page-level order sequence as example. We combine the representation of target page with the page-level representations of order sequence to form the input matrix $\mathbf{B}_\text{o} = \{\mathbf{h}_{t},\mathbf{h}^\text{o}_{1},\cdots,\mathbf{h}^\text{o}_{N_\text{o}}\}$. 
Similar to \citet{xie2021deep}, we then use multi-head self-attention to generate the interacted page-level order sequence representation and formulate the result as $\mathbf{z}^\text{o}$.

Notice that, the length of page-level implicit feedback (i.e., pull-down) sequence is obviously longer than the other three, which can be noisy to some extent \cite{xie2021deep}, since the items exposed are carefully selected by ranking strategies or user may scroll too fast to notice these items. 
Accordingly, we use the other three sequences information to discern the page-level arrangement that user may or may not perfer in the page-level pull-down feedback sequence. Take $\mathbf{z}^\text{o}$ as example, the denoising representation of page-level pull-down feedback $\mathbf{z}^\text{p(o)}$ is calculated as: 
\begin{equation}
  \begin{aligned}
    w_{p_i}^{\text{o}} =\ \text{MLP}_2 &\Big(\mathbf{h}_i^\text{p}||\mathbf{z}^\text{o}||(\mathbf{h}_i^\text{p} \odot \mathbf{z}^\text{o})||(\mathbf{h}_i^\text{p}-\mathbf{z}^\text{o}) \Big), \\
    w_{p_i}^{\text{o}} &=  \frac{\exp (w_{p_i}^{\text{o}})}{\sum_{j=1}^{N^\text{p}}\exp (w_{p_j}^{\text{o}})}, \\
    \mathbf{z}^\text{p(o)} & =\ \sum_{i=1}^{N_{\text{p}}} w_{p_i}^{\text{o}} \mathbf{h}_i^\text{p}.
  \end{aligned}
\end{equation}

We concatenate the extracted representations as the page-level feedback representation in current channel. The outputs of different channels will be concatenated together as follows:
\begin{equation}
  \begin{aligned}
    \mathbf{c}_i =\   &\mathbf{z}^\text{o} || \mathbf{z}^\text{c} || \mathbf{z}^\text{p} || \mathbf{z}^\text{l} || \mathbf{z}^\text{p(o)} || \mathbf{z}^\text{p(c)} || \mathbf{z}^\text{p(l)}|| \mathbf{h}^\text{t},
    \\
    & \mathbf{e}_\text{MCIM} =\ \mathbf{c}_1 || \mathbf{c}_2 || ... || \mathbf{c}_T.
  \end{aligned}
\end{equation}

\subsection{Optimization Objective}
We concatenate the output of MCIM with the embeddings of context and user profile to predict the value $Q$ through an MLP:
\begin{equation}
  \begin{aligned}
    Q(s,a) = \text{MLP}_3\Big(\mathbf{e}_\text{MCIM} || \mathbf{e}_c || \mathbf{e}_u \Big).
  \end{aligned}
\end{equation}

For each iteration, we sample a batch of transitions $B$ from the offline dataset and update the agent using gradient back-propagation w.r.t. the loss \cite{mnih2015human}:
\begin{equation}
\label{eq:loss}
L(B) = \frac{1}{|B|}\sum_{(s,a,r,s')\in B} \Big( r + \gamma \max_{a'\in\mathcal{A}} Q(s', a') - Q(s,a) \Big) ^ 2.
\end{equation}

\section{Experiments}
We will evaluate our DPIN through offline and online experiments in this section. 
In offline experiments, we compare our method with existing state-of-the-art baselines and analyze the role of different units and different page-level behavior sequences.
In online experiments, we compare our method with the previous strategy deployed on Meituan food delivery platform using an online A/B test.
\subsection{Experimental Settings}
\subsubsection{Dataset}
We collect the dataset by running an exploratory policy on Meituan food delivery platform during January 2022. The dataset contains 12,411,532 requests, 1,732,492 users, 358,394 ads and 710,937 organic items. We use the user's request within 30 days to obtain the user's four types of page-level sequences (page-level order, click, pull-down and leave). The average length of the four sequences is $4.63$, $10.10$, $39.4$ and $12.68$, respectively.

\subsubsection{Evaluation Metrics}

We evaluate with the ads revenue $R^\text{ad}=\sum r^\text{ad}$, the service fee $R^\text{fee}=\sum r^\text{fee}$. See the definition in Section \ref{sec:problem}.
\subsubsection{Hyperparameters}
We apply a gird search for the hyperparameters. 
The length of each sequence is truncated (or padded) to $10$, the number of channel is $5$, the hidden layer sizes of all MLPs are $(128, 64, 32)$, the $\tau$ is $0.9$, the learning rate is $10^{-3}$, the optimizer is Adam \cite{kingma2014adam} and the batch size is 8,192.

\subsection{Offline Experiment}
In this section, we train our method with offline data and evaluate the performance using an offline estimator. 
Through extended engineering, the offline estimator models the user preference and aligns well with the online service.

\subsubsection{Baselines}
We compare our method with the following representative RL-based dynamic ads slots methods:
\begin{itemize}[leftmargin=*]
   \item \textbf{HRL-Rec} divides the integrated recommendation into two levels of tasks and solves using hierarchical reinforcement learning. 
  \item \textbf{DEAR} designs a deep Q-network architecture to determine three related tasks jointly, i.e., i) whether to insert an ad to the recommendation list, and if yes, ii) the optimal ad and iii) the optimal location to insert.
  \item \textbf{CrossDQN} takes the crossed state-action pair as input and allocates slots in one page at a time. It designs some units (e.g., MCAU) to optimize the combinatorial impact of the items on user behavior.
  \item \textbf{CrossDQN \& DIN} introduces point-level order sequence into CrossDQN . The sequence is modeled with DIN \cite{zhou2018DIN}. 
  \item \textbf{CrossDQN \& DFN} introduces point-level order, click, pull-down, leave sequences into CrossDQN . The four sequences are modeled with DFN \cite{xie2021deep}.
  \item \textbf{CrossDQN \& RACP} introduces page-level order, click, pull-down, leave sequences into CrossDQN. The sequence is modeled with RACP \cite{xie2021deep}.
\end{itemize}

\subsubsection{Performance Comparison}
We present the experimental results in Table \ref{result}. Compared with all these baselines, our method achieves strongly competitive performance on both the ads revenue and the service fee. Specifically, our method improves over the best baseline w.r.t. $R^\text{ad}$ and $R^\text{fee}$ by 1.7\% and 2.2\% separately. The superior performance of our method justifies that effectiveness of modeling page-level user preference through multiple types of page-level behavior sequences.

\subsubsection{Ablation Study}
To verify the impact of our designs, we study six ablated variants of our method  and have the following findings:
i) The performance gap between w/ and w/o CL verifies the effectiveness of modeling user perference for item arrangement under different receptive fields. ii) The performance gap between w/ and w/o IPIU verifies the effectiveness of calculating the correlation between target page and page-level sequences. iii) The performance gap between w/ and w/o $\{\mathbf{E}^{\text{p}}\}\& \{\mathbf{E}^{\text{l}}\}$ and the performance gap w/ and w/o $\{\mathbf{E}^{\text{p}}\}\& \{\mathbf{E}^{\text{l}}\}\& \{\mathbf{E}^{\text{c}}\}$ verify the effectiveness of utilizing multiple types of page-level feedback.

\subsubsection{Hyperparameter Analysis}
We analyze the sensitivity of the number of channels in our method. The experimental results\footnote{The specific experimental results are not presented due to space reasons.} show that model achieve better performance as the number of channels increases, especially the number of channels grows from 1 to 3.

\subsection{Online Results}
We compare DPIN with CrossDQN and both strategies are deployed on Meituan food delivery platform through online A/B test. As a result, we find that $R^\text{ad}$ and $R^\text{fee}$ increase by 1.5\% and 1.7\%, which demonstrates that DPIN can greatly increase the platform revenue.

\begin{table}[!tb]
  \caption{The experimental results. Each experiment is presented in the form of mean $\pm$ standard deviation. The improvement means the improvements of our method across the best baselines.}
  \centering
  \renewcommand\arraystretch{1.15}
  \setlength{\tabcolsep}{1.3mm}
  \begin{tabular*}{1\linewidth}{l|c|c}
  \hline
  \rule{0pt}{11.4pt}
      model & $R^{\text{ad}}$ & $R^{\text{fee}}$ \\ \hline  \hline 
      HRL-Rec & 0.1114\ ($\pm$0.0002) & 0.9485\ ($\pm$0.0255) \\ \hline
      DEAR & 0.1119\ ($\pm$0.0003) & 0.9545\ ($\pm$0.0198) \\ \hline
      CrossDQN & 0.1149\ ($\pm$0.0005) & 0.9761\ ($\pm$0.0063) \\ \hline
      CrossDQN\&DIN & 0.1150\ ($\pm$0.0006) & 0.9789\ ($\pm$0.0082) \\ \hline
      CrossDQN\&DFN & 0.1153\ ($\pm$0.0003) & 0.9824\ ($\pm$0.0050) \\ \hline
      CrossDQN\&RACP & 0.1157\ ($\pm$0.0003) & 0.9836\ ($\pm$0.0100) \\ \hline \hline 
      \textbf{Our method} & \textbf{0.1181\ ($\pm$0.0003)} & \textbf{1.0105\ ($\pm$0.0102)} \\ \hline
      \ - w/o $\text{CL}$ & 0.1161\ ($\pm$0.0007) & 0.9883\ ($\pm$0.0033 \\ \hline
      \ - w/o $\text{IPAU}$ & 0.1167\ ($\pm$0.0005) & 0.9872\ ($\pm$0.0098) \\ \hline
      \ - w/o $\text{IPIU}$ & 0.1160\ ($\pm$0.0002) & 0.9843\ ($\pm$0.0067) \\ \hline
      \ - w/o $\text{MCIM}$ & 0.1151\ ($\pm$0.0006) & 0.9781\ ($\pm$0.0059) \\ \hline
      \ - w/o $ \{\mathbf{E}^{\text{p}}\}\& \{\mathbf{E}^{\text{l}}\}$ & 0.1163\ ($\pm$0.0005) & 0.9999\ ($\pm$0.0096) \\ \hline
      \ - w/o $\{\mathbf{E}^{\text{p}}\}\& \{\mathbf{E}^{\text{l}}\}\& \{\mathbf{E}^{\text{c}}\}$ & 0.1158\ ($\pm$0.0003) & 0.9957\ ($\pm$0.0189) \\ \hline
       \hline 
      Improvement & 1.7\% & 2.2\% \\ \hline
  \end{tabular*}
  \label{result}
\end{table}

\section{Conclusions}
In this paper, we propose an method for page-level historical behavior sequence modeling on ads allocation problem. 
Specifically, we introduce four different types of page-level feedback (i.e., page-level order, click, pull-down, leave) as input, and capture user perference for item arrangement under different receptive fields through the multi-channel interaction module. 
Practically, both offline experiments and online A/B test have demonstrated the superior performance and efficiency of our method.

\balance
\bibliographystyle{ACM-Reference-Format}
\bibliography{pagewise}

\appendix

\end{document}